\documentclass[10pt,twocolumn,letterpaper]{article}

\usepackage[pagenumbers]{cvpr} 
\usepackage{algorithmic}
\usepackage{algorithm}
\usepackage{graphicx}
\usepackage[dvipsnames]{xcolor}

\definecolor{cvprblue}{rgb}{0.21,0.49,0.74}
\usepackage[pagebackref,breaklinks,colorlinks,citecolor=cvprblue]{hyperref}

\def\paperID{2098} 
\def\confName{CVPR}
\def\confYear{2024}

\title{Time-Efficient Light-Field Acquisition Using Coded Aperture and Events}

\author{Shuji Habuchi$^\dagger$\hspace{5mm}
Keita Takahashi$^\dagger$\hspace{5mm}
Chihiro Tsutake$^\dagger$\hspace{5mm}
Toshiaki Fujii$^\dagger$\hspace{5mm}
Hajime Nagahara$^\ddagger$ \\
$^\dagger$ Nagoya University, Japan\hspace{5mm}
$^\ddagger$ Osaka University, Japan
}

\begin{document}
\maketitle
{
\begin{abstract}
We propose a computational imaging method for time-efficient light-field acquisition that combines a coded aperture with an event-based camera. Different from the conventional coded-aperture imaging method, our method applies a sequence of coding patterns during a single exposure for an image frame. The parallax information, which is related to the differences in coding patterns, is recorded as events. The image frame and events, all of which are measured in a single exposure, are jointly used to computationally reconstruct a light field. We also designed an algorithm pipeline for our method that is end-to-end trainable on the basis of deep optics and compatible with real camera hardware. We experimentally showed that our method can achieve more accurate reconstruction than several other imaging methods with a single exposure. We also developed a hardware prototype with the potential to complete the measurement on the camera within 22 msec and demonstrated that light fields from real 3-D scenes can be obtained with convincing visual quality. Our software and supplementary video are available from our project website\footnote{https://www.fujii.nuee.nagoya-u.ac.jp/Research/EventLF/}.
\end{abstract}

\begin{figure*}[t]
\centering
\includegraphics[width=.95\linewidth]{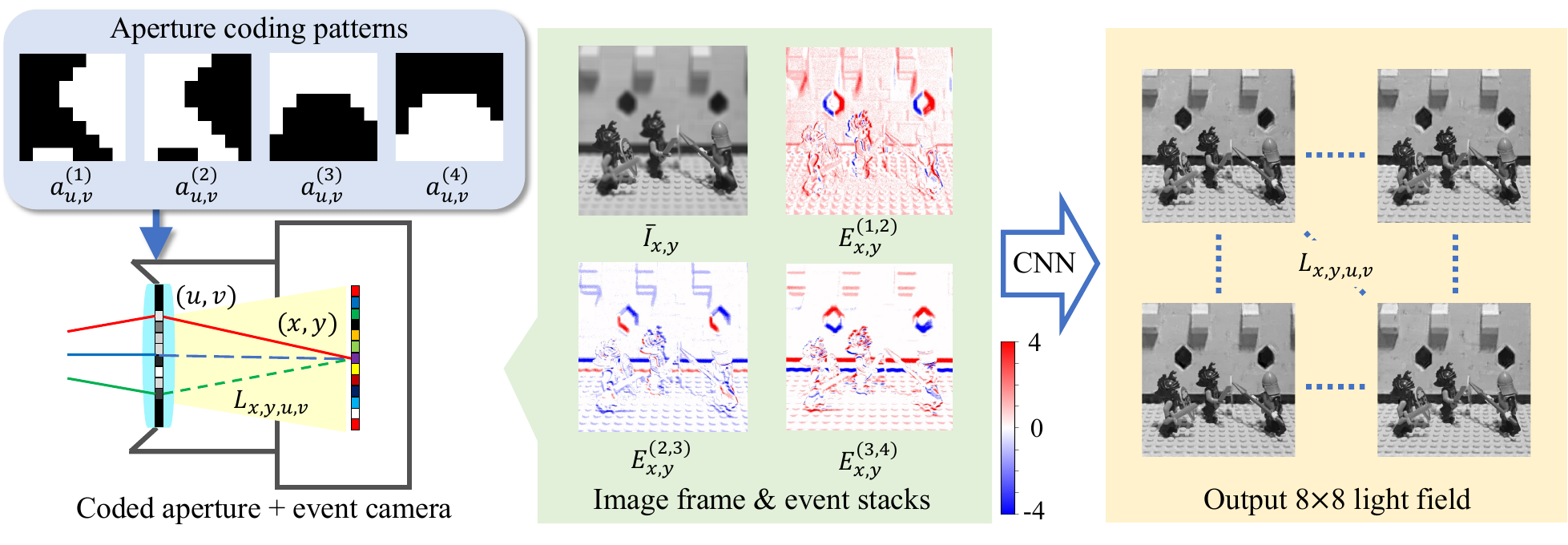}
\caption{Overview of our method. Four coding patterns are applied to aperture plane during single exposure of image frame. Image and events are jointly used to reconstruct light field through convolutional neural network (CNN).}
\label{fig:overview}
\end{figure*}

\section{Introduction}

A light field is usually represented as a set of multi-view images that captures a target 3-D scene from a dense 2-D grid of viewpoints. Light fields have been used for various applications such as depth estimation~\cite{honauer2017dataset,shin18epinet,khan2021edgeaware}, object/material recognition~\cite{Maeno2013,wang20164d}, view synthesis~\cite{Kalantari2016,mildenhall2019llff, broxton2020immersive}, and 3-D display~\cite{wetzstein2012tensor,huang2015light,lee2016additive}. In this paper, we consider light-field acquisition from static 3-D scenes.

Due to the large number of images contained in a light field (e.g., $8 \times 8$ views), how to acquire a light field has been a long-standing issue~\cite{wilburn2005high,ng2006digital,fujii2006multipoint,Taguchi2009}. Since views included in a light field are highly redundant with each other, view-by-view sampling seems to be a waste of resources. To achieve more efficient acquisition, researchers investigated coded-aperture imaging~\cite{liang2008programmable,nagahara2010programmable,Inagaki_2018_ECCV,Sakai_2020_ECCV,Guo2022TPAMI}, with which semi-transparent coding patterns are placed at the aperture plane of a camera. With this method, the light field of a target scene is optically encoded before being recorded on the image sensor. Some images taken from a stationary camera with different coding patterns are used to computationally reconstruct the original light field. As indicated from several studies~\cite{Inagaki_2018_ECCV,Guo2022TPAMI,Tateishi_2022_IEICE}, the number of images to acquire can be reduced to only a few (e.g., 2--4). 

An issue with coded-aperture imaging is the long measurement time, since two or more images should be acquired in sequence. As a solution for this issue, we propose a time-efficient light-field acquisition method that can complete the measurement \textit{in a single exposure}. Our method combines a coded aperture~\cite{liang2008programmable,nagahara2010programmable,Inagaki_2018_ECCV,Sakai_2020_ECCV,Guo2022TPAMI} and an event-based camera~\cite{eventcamerasurvey2022,Brandli_2014_IEEE}, as shown in Fig.~\ref{fig:overview}, that can simultaneously capture image frames and events. Our method is based on the premise that the aperture coding can be controlled faster than the frame-rate of the image sensor.\footnote{Some electronic devices used for implementing a coded aperture can run much faster than standard image sensors.} We apply several coding patterns in sequence during a single exposure for an image frame. Therefore, we do not directly observe individual coded-aperture images, but we have the sum of them as a single image frame. The camera also measures the events asynchronously during the exposure. Since the target scene is assumed static, the events are caused exclusively by the change in coding patterns. In other words, we actively induce the events by changing the coding patterns over time. These events include the information related to the parallax among different viewpoints, which is essential for light-field/3-D reconstruction. The image frame and events, all of which are measured in a single exposure, are jointly used to computationally reconstruct the light field. 

We first formalize our imaging method and clarify the \textit{quasi-equivalence} between our method and the baseline coded-aperture imaging method. We then discuss our design of an end-to-end trainable algorithm tailored for our imaging method while considering the compatibility with real camera hardware. To validate our method, we conducted quantitative evaluations, in which the imaging process was computationally simulated, and real-world experiments using our prototype camera to capture real 3-D scenes. Experimental results show that our method can reconstruct light fields more accurately than several other imaging methods with a single exposure, and it works successfully with our prototype camera in capturing real 3-D scenes with convincing visual quality.

To the best of our knowledge, we are the first to investigate the combination of coded-aperture imaging and events in the context of computational light-field imaging. Our contribution is not limited to shortening the measurement time for a light field, but it enables us to \textit{go beyond the limitation of the frame-rate} of image sensors; Our method can {better utilize the time resource during a single exposure}, and obtain {more information per unit time (i.e., being time-efficient)} than the baseline coded-aperture imaging method. Our method is also distinctive in the sense that events are induced actively by the camera's optics rather than the moving objects or ego-motion of the camera.

\section{Related Works}

The most straight-forward approach to light-field acquisition is to construct an array of cameras~\cite{wilburn2005high,fujii2006multipoint,Taguchi2009}, which involves costly and bulky hardware. Lens-array-based cameras~\cite{adelson1992single,arai1998gradient,ng2005light,ng2006digital} gained popularity because a light field can be captured in a single shot. However, this type of camera has an intrinsic trade-off between the number of views and the spatial resolution of each view. Mask-based coded-imaging methods~\cite{veeraraghavan2007dappled,liang2008programmable,nagahara2010programmable,marwah2013compressive,Inagaki_2018_ECCV,nabati2018colorLF,vadathya2019,Guo2022TPAMI} have been developed to increase the efficiency of light-field acquisition. With coded-aperture imaging, two to four images, taken from a stationary camera with different aperture-coding patterns, are sufficient to computationally reconstruct a light field with $8 \times 8$ views in full-sensor resolution~\cite{Inagaki_2018_ECCV,Guo2022TPAMI,Tateishi_2022_IEICE}. However, since several coded images need to be acquired in sequence, the lengthy measurement time remains an issue. 
Joint aperture-exposure coding~\cite{Tateishi_2021_ICIP,vargas_2021_ICCV,Mizuno_2022_CVPR,Tateishi_2022_IEICE} enables more flexible coding patterns during a single exposure but comes with complicated hardware implementation; as far as we know, only Mizuno et al.~\cite{Mizuno_2022_CVPR} reported a working prototype for this method but with awkward hardware restrictions for the non-commercialized image sensor. Our method also applies several aperture-coding patterns during a single exposure, but we combine them with an off-the-shelf event camera to achieve time-efficient and accurate light-field acquisition.

Event cameras~\cite{eventcamerasurvey2022,Brandli_2014_IEEE} are bio-inspired sensors that can record intensity changes asynchronously at each pixel with a very low latency. Compared with ordinary frame-based cameras, event cameras can capture more fine-grained temporal information in a higher dynamic range, which opens up many applications such as optical-flow estimation, deblurring, video interpolation, and camera pose estimation. Event cameras have also been used extensively for 3-D reconstruction~\cite{Kim2016Real,Rebecq2018EMVS,Zhou2018Semi,Zhou2021Event,rudnev2023eventnerf,Ma_2023_ICCV}. In these studies, however, the events were usually caused either by the moving objects or ego-motion of the camera. Our method can be regarded as a new application of an event camera; the events are induced actively by the camera's optics in the framework of computational imaging. 

Single-view view synthesis (SVVS)~\cite{Niklaus_TOG_2019,Chen2020LFGAN, single_view_mpi,VariableMPI2020ACM,Trevithick_2021_ICCV,yu2021pixelnerf,mine2021ICCV,Chan_2021_CVPR,Xu_2022_SinNeRF,Chan_2023_ICCV} is used to reconstruct a 3-D scene from a single image. Since this is geometrically an ill-posed problem, SVVS methods rely on implicit prior knowledge learned from the training dataset rather than physical cues. These methods are not necessarily designed to be physically accurate but to hallucinate a visually-plausible 3-D scene. Our method takes an orthogonal approach to SVVS; we use not only a single image but also a coded aperture and events to obtain solid physical cues from the target 3D scene.

Various computational imaging methods have been developed on the basis of deep-optics~\cite{Iliadis_2016_mask,Nie_2018_CVPR, Yoshida_2018_ECCV,Inagaki_2018_ECCV,Wu_2019_ICCP,Li_2020_ICCP,Mizuno_2022_CVPR,Wang_2023_ICCV}, with which the camera-side optical elements and the computational algorithm are jointly optimized in a deep-learning-based framework. Our work is pioneering in applying deep-optics to event cameras.

\section{Proposed Method}

\subsection{Background and Basics}

A schematic diagram of our camera is shown in Fig.~\ref{fig:overview} (left). All the light rays coming into the camera can be parameterized by $(x,y,u,v)$, where $(u,v)$ and $(x,y)$ denote the positions on the aperture and imaging planes, respectively. Therefore, a light field $L$ is defined over $(x,y,u,v)$. We assume that each light ray has only a monochrome intensity; but the extension to RGB color is straight-forward in theory. We also assume that $x,y,u,v$ take discretized integer values; thus, $L$ is equivalent to a set of multi-view images, where $(x,y)$ and $(u,v)$ respectively denote the pixel position and viewpoint. The arrangement of the viewpoints is assumed to be $8 \times 8$ ($ u,v \in \{1, \dots, 8\}$). Each element of $L$ is described using subscripts as $L_{x,y,u,v}$. The goal of our method is to reconstruct $L$ of the target scene from the data measured on the camera in a single exposure. 

Before describing our imaging method in Section 3.2, we mention two previous imaging methods that use \textit{a coded aperture with a frame-based camera} (no events available). 

\textbf{Coded-aperture imaging}~\cite{liang2008programmable,nagahara2010programmable,Inagaki_2018_ECCV,Sakai_2020_ECCV,Guo2022TPAMI}: 
To optically encode a light field $L$ along the viewpoint dimension, a sequence of light-attenuating patterns, $a^{(1)}, \dots, a^{(N)}$, where $a^{(n)}_{u,v} \in [0, 1]$ and $u,v \in \{1, \dots, 8\}$, is placed at the aperture plane. Each pixel under the $n$-th coding pattern is described as the weighted sum of the light rays over $(u,v)$:
\begin{align}
I^{(n)}_{x,y} = \sum_{u,v} a^{(n)}_{u,v}L_{x,y,u,v}.
\label{eq:coded-aperture}
\end{align}
The light field $L$ is computationally reconstructed from $N$ images taken with different coding patterns, $I^{(1)}, \dots, I^{(N)}$. The images under different coding patterns would have parallax with each other, which is essential for 3-D/light-field reconstruction. As demonstrated in previous studies~\cite{Inagaki_2018_ECCV,Guo2022TPAMI,Tateishi_2022_IEICE}, a relatively small $N$ (e.g. $N=4$) is sufficient for accurate reconstruction. However, since $N$ (more than one) images should be acquired in sequence, the lengthy measurement time remains an issue.

\textbf{Joint aperture-exposure coding}~\cite{Tateishi_2021_ICIP,vargas_2021_ICCV,Mizuno_2022_CVPR,Tateishi_2022_IEICE}: This is an advanced form of coded-aperture imaging; $N$ coding patterns are applied to both the aperture and imaging planes synchronously \textit{during} a single exposure. The coding patterns for the imaging plane are described as $p^{(1)}, \dots, p^{(N)}$, where $p^{(n)}_{x,y} \in \{0, 1\}$. The imaging process is described as
\begin{align}
I_{x,y} &= \sum_n \sum_{u,v} a^{(n)}_{u,v} \: p^{(n)}_{x,y}  \: L_{x,y,u,v}.
\label{eq:joint-coding}
\end{align}
While the light field $L$ can be reconstructed from a single observed image alone, the increased complexity of the coding scheme (Eq.~(\ref{eq:joint-coding})) makes its hardware implementation very difficult.

\subsection{Combining Coded Aperture and Events}
\label{sec:theory}


As shown in Fig.~\ref{fig:overview}, our method uses \textit{a coded aperture with an event camera} that can obtain both the image frames and events simultaneously (e.g., DAVIS 346 camera). Similar to the baseline coded-aperture imaging method (Eq.~(\ref{eq:coded-aperture})), we apply $N$ aperture-coding patterns ($a^{(1)}, \dots, a^{(N)}$) in sequence. However, we do so \textit{in a single exposure} for an image frame. Therefore, we do not directly observe individual coded-aperture images, $I^{(1)}, \dots, I^{(N)}$, but we have their sum as the image frame:
\begin{align}
\bar{I}_{x,y} = \sum_{n} I^{(n)}_{x,y}.
\label{eq:APS}
\end{align}
The camera also measures the events at each pixel $(x,y)$ asynchronously during the exposure. We denote an event occurring at time $t$ as $e_{x,y,t}\in \{+1, -1\}$, where the positive/negative signs correspond to the increase/decrease in the intensity at pixel $(x,y)$. Since the target scene is assumed static, the events are caused exclusively by the change in the coding patterns. Although the pattern changes instantly, the camera responds gradually in a very short but no-zero transient time. We denote the transient time between $a^{(n)}$ and $a^{(n+1)}$ as $T^{(n, n+1)}$. We sum the events during the transient time to obtain an event stack as
\begin{align}
E^{(n,n+1)}_{x,y} =\sum_{t\: \in \: T^{(n, n+1)}} e_{x,y,t}.
\label{eq:EventStack}
\end{align}
As shown in Fig.~\ref{fig:overview}, $E^{(n,n+1)}$ includes the parallax information between $I^{(n)}$ and $I^{(n+1)}$, which is essential for 3-D reconstruction. Our goal is to reconstruct the original light field $L$ from a single image frame $\bar{I}$ and ($N-1$) event stacks, $E^{(1,2)},\dots, E^{(N-1,N)}$, all of which are measured during a single exposure. 

\textbf{Quasi-equivalence}. We theoretically relate our imaging method to the baseline coded-aperture imaging method. The camera records an event at time $t$ when the intensity change (in log scale) exceeds a contrast threshold $\tau$ as
\begin{align}
|\log(I_{x,y,t}) - \log(I_{x,y,t_0})| > \tau
\label{eq:event-model}
\end{align}
where $t_0$ is the time when the previous event was recorded. On the basis of Eq.~(\ref{eq:event-model}), we derive an approximate relation between an event stack (Eq.~(\ref{eq:EventStack})) and the two consecutive coded-aperture images (Eq.~(\ref{eq:coded-aperture})) as 
\begin{align}
E^{(n,n+1)}_{x,y} \approx \frac{\log\left(I^{(n+1)}_{x,y}\right) - \log\left(I^{(n)}_{x,y}\right)}{\tau}.
\label{eq:event-image}
\end{align}
Assuming that equality holds for Eq.~(\ref{eq:event-image}) and combining it with Eq.~(\ref{eq:APS}), we can obtain $I^{(1)}, \dots, I^{(N)}$ as
\begin{align}
I^{(n)}_{x,y} &=  I^{(1)}_{x,y} \exp\left(\tau \textstyle{\sum_{2 \le k \le n}}  E^{(k-1,k)}_{x,y}\right)  (n\ge 2) \label{eq:solve} \\
I^{(1)}_{x,y} & = \dfrac{\bar{I}_{x,y}}{1 + \sum_{2 \le n \le N} \exp\left(\tau \sum_{2\le k \le n}  E^{(k-1,k)}_{x,y}\right)}. \label{eq:sigma}
\end{align}
This means that under the equality assumption for Eq.~(\ref{eq:event-image}), $N$ coded-aperture images ($I^{(1)}, \dots, I^{(N)}$) can be derived analytically from the data observed with our imaging method ($\bar{I}$ and $E^{(1,2)}, \dots, E^{(N-1,N)}$). Therefore, we can state that our imaging method is \textit{quasi-equivalent} to the baseline coded-aperture imaging method. 

Interestingly, this type of quasi-equivalence does not hold for joint aperture-exposure coding. By using Eq.~(\ref{eq:coded-aperture}), the imaging process of Eq.~(\ref{eq:joint-coding}) is rewritten as
\begin{align}
I_{x,y} = \sum_n p^{(n)}_{x,y} \: I^{(n)}_{x,y}.
\label{eq:joint-coding2}
\end{align}
Obviously, it is impossible to analytically obtain $N$ coded-aperture images ($I^{(1)}, \dots, I^{(N)}$) from a single observed image $I$ alone. Therefore, our imaging method has a theoretical advantage over joint aperture-exposure coding. However, this theory alone is insufficient to ensure the practicality of our method. The actual event data are very noisy and harshly quantized (by the contrast threshold $\tau$), which breaks the equality assumption for Eq.~(\ref{eq:event-image}).

\subsection{Algorithm}

We developed an end-to-end trainable algorithm on the basis of deep-optics~\cite{Iliadis_2016_mask,Nie_2018_CVPR, Yoshida_2018_ECCV,Inagaki_2018_ECCV,Wu_2019_ICCP,Li_2020_ICCP,Mizuno_2022_CVPR,Wang_2023_ICCV}, in which the camera-side optical-coding patterns and the light-field reconstruction algorithm were jointly optimized in a deep-learning-based framework. We carefully designed each part of our algorithm to ensure the compatibility with real camera hardware. Although we specifically mention the hardware setup that is available to us, the ideas behind our design would be useful for other possible hardware setups. We set $N=4$  unless otherwise mentioned.

Our algorithm consists of two parts: AcqNet and RecNet. AcqNet describes the data-acquisition process using a coded aperture and an event camera as
\begin{align}
\bar{I}, E^{(1,2)}, E^{(2,3)},E^{(3,4)} = \mbox{AcqNet}(L).
\end{align}
The trainable parameters of AcqNet are related to the aperture's coding patterns. RecNet receives the observed data as the input and reconstructs the original light field as
\begin{align}
\hat{L} = \mbox{RecNet}(\bar{I}, E^{(1,2)}, E^{(2,3)},E^{(3,4)} ).
\end{align}
AcqNet and RecNet are jointly trained to minimize the reconstruction (MSE) loss between $L$ and $\hat{L}$. Once the training is finished, AcqNet is replaced with the physical imaging process of the camera hardware, in which the coding patterns are adjusted to the learned parameters of AcqNet. The data acquired from the camera are fed to RecNet to reconstruct the light field of a real 3-D scene.

\textbf{Hardware-driven constraints for coded aperture}. Similar to some previous studies~\cite{nagahara2010programmable,marwah2013compressive,Inagaki_2018_ECCV}, we used a liquid-crystal-on-silicon (LCoS) display (Forth Dimension Displays, SXGA-3DM)  to implement a coded aperture. This display can output a sequence of semi-transparent coding patterns repeatedly. We need to consider the following two constraints. \textbf{Binary constraint}. Although our LCoS display can support both binary and grayscale patterns, a grayscale pattern is actually represented as a temporal series of multiple binary patterns; a grayscale transmittance is represented as the ratio of 0/1 periods. To avoid unintended 0/1 flips, we choose to use only binary patterns for aperture coding. \textbf{Complementary constraint}. Our LCoS display requires a ``DC balance''; a certain pattern $a$ and its complement $a^{\ast} = 1-a$ should be included in the sequence. Since the events are recorded continuously over time, we use both $a$ and $a^{\ast}$ as the coding patterns.

\begin{algorithm}[tb]
    \caption{Pseudo-code for AcqNet}
    \begin{algorithmic}[1] 
    \STATE trainable tensors: $\alpha, \beta \in {\cal R}^{8 \times 8} $  
    \STATE forward($L$): 
    \STATE \hspace{\algorithmicindent} set $s$  
    \STATE \hspace{\algorithmicindent} $a^{(1)},\: a^{(3)}= \mbox{sigmoid}(s\alpha), \mbox{sigmoid}(s\beta)$ 
    \STATE \hspace{\algorithmicindent} $a^{(2)}, \: a^{(4)} = 1-a^{(1)}, \: 1-a^{(3)}$  
    \STATE \hspace{\algorithmicindent} \textbf{for} $n$ in [1, 2, 3, 4]:
    \STATE \hspace{2\algorithmicindent} compute $I^{(n)}$ by Eq.~(\ref{eq:coded-aperture}) 
    \STATE \hspace{2\algorithmicindent} \textbf{if} $n>1$: 
    \STATE \hspace{3\algorithmicindent} compute $E^{(n-1,n)}$ by Eq.~(\ref{eq:ESIM})  
    \STATE \hspace{2\algorithmicindent} \textbf{end}
    \STATE \hspace{\algorithmicindent} \textbf{end}
    \STATE \hspace{\algorithmicindent} compute $\bar{I}$ by Eq.~(\ref{eq:APS})  
    \STATE \hspace{\algorithmicindent} return $\bar{I}$, $E^{(1,2)}$, $E^{(2,3)}$, $E^{(3,4)}$
    \end{algorithmic}
\end{algorithm}

\textbf{AcqNet}. A pseudo-code of AcqNet is presented in Algorithm 1. In lines 4 and 5, we make the coding patterns compatible with the binary and complementary constraints. More specifically, we prepare two sets of trainable tensors, each with $8\times8$ elements, denoted as $\alpha$ and $\beta$. They are multiplied by the scale parameter $s$ then fed to the sigmoid function to produce the coding patterns $a^{(1)}$ and $a^{(3)}$. As the training proceeds, $s$ gradually increases, which forces $a^{(1)}$ and $a^{(3)}$ to gradually converge to binary patterns. Moreover, $a^{(2)}$ and $a^{(4)}$ are made to be complementary to $a^{(1)}$ and $a^{(3)}$, respectively. In line 9, we compute an event stack from the consecutive coded-aperture images as
\begin{align}
\!\! E^{(n,n+1)}_{x,y} = Q\!\left( \frac{\log(I^{(n+1)}_{x,y} + \epsilon) - \log(I^{(n)}_{x,y} + \epsilon)}{\tau + n_{x,y}} \right)
\label{eq:ESIM}
\end{align}
where the intensities of $I^{(n)}$ and $I^{(n+1)}$ are normalized to $[0,1]$, $Q$ is a quantization operator and defined as $Q(x) = \mbox{sign}(x)\mbox{floor}(|x|)$, and $\epsilon = 0.01$. We add a zero-means Gaussian noise $n$ with $\sigma=0.021$ to $\tau$ to account for the randomness of the sensor. The formulation of Eq.~(\ref{eq:ESIM}) follows a widely used event simulator~\cite{Rebecq18corl}, but we implement it as being differentiable. Following previous studies~\cite{Inagaki_2018_ECCV,Tateishi_2022_IEICE,Mizuno_2022_CVPR}, we also add a zero-means Gaussian noise ($\sigma=0.005$ w.r.t. the normalized intensity range $[0,1]$) to the image frame $\bar{I}$ to account for the measurement noise.

A challenging issue with Eq.~(\ref{eq:ESIM}) is how to determine the contrast threshold $\tau$. When the imaging process is computationally simulated, a smaller $\tau$ is better. This is because a smaller $\tau$ leads to a more fine-grained observation (a larger number of events), which in turn results in more accurate light-field reconstruction. However, it is difficult to determine a specific $\tau$ that is compatible with a real event camera (e.g., a DAVIS 346 camera) the $\tau$ of which cannot be controlled in an explicit (direct) manner. With our experimental configuration, we empirically estimate $\tau \simeq 0.15$, but it depends on the configuration. To address this issue, we treat $\tau$ as being variable, which makes the model of Eq.~(\ref{eq:ESIM}) more flexible. Specifically, we randomly draw $\tau$ from $[0.075, 0.3]$ for each batch during training, which enables the algorithm to be independent of a specific $\tau$.

\begin{figure}[t]
\centering
\includegraphics[width=.99\linewidth]{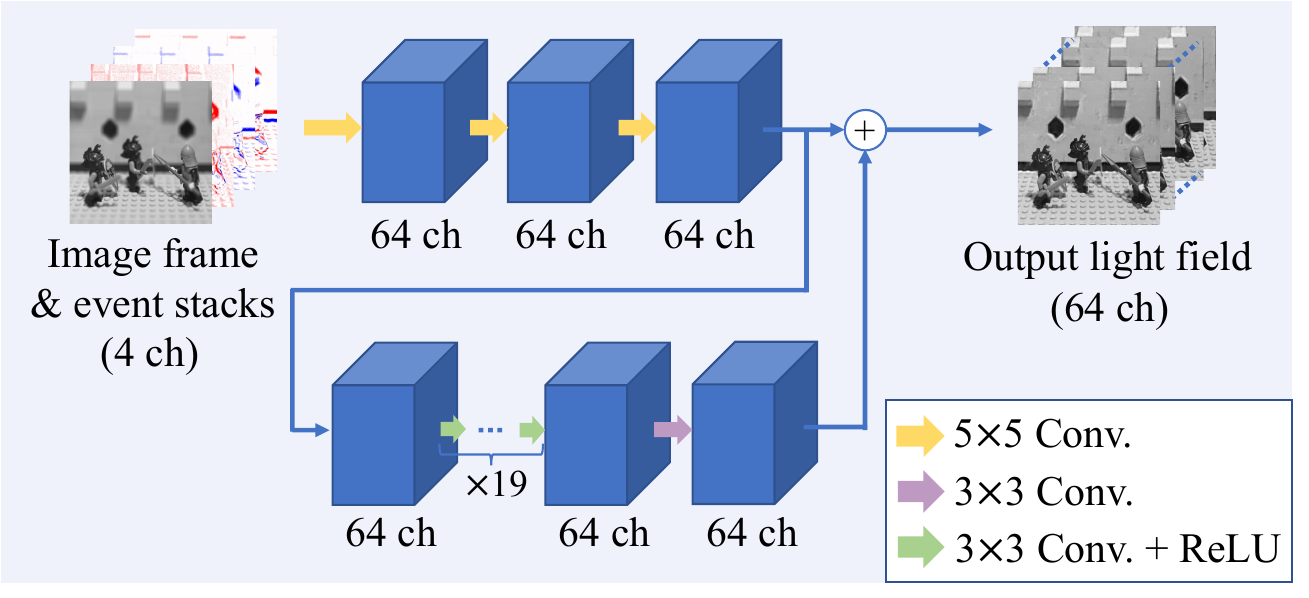}
\caption{Network architecture of RecNet}
\label{fig:network}
\end{figure}

\textbf{RecNet}. Since network architectures are not the main focus of this paper, we used a plain network architecture consisting of a sequence of 2-D convolutional layers. The data obtained from the camera ($\bar{I}, E^{(1,2)}, E^{(2,3)}$, and $ E^{(3,4)}$ stacked along the channel dimension) are fed to RecNet; the output from RecNet is a tensor with 64 channels corresponding to 64 views of the reconstructed light field (see Fig.~\ref{fig:network} for more details). Similar architectures were used in previous works~\cite{Inagaki_2018_ECCV,Tateishi_2022_IEICE}, achieving a good balance between the reconstruction quality and computational cost. Our plain architecture can be used for other imaging methods with minimal modifications (only by changing the number of channels for the input layer), which makes the comparison easier. Exploration for better network architectures is left as future work.

\textbf{Training and loss function}. AcqNet and RecNet are implemented using PyTorch 2.0 and jointly trained on the BasicLFSR dataset~\cite{BasicLFSR}. We extract 29,327 training samples, each with $64 \times 64$ pixels and $8\times8$ views, from 144 light fields designated for training. The disparities among the neighboring viewpoints are mostly limited within $[-3,3]$ pixels. This is suitable for our method because the 8$\times$8 viewpoints are arranged on the small aperture plane of the camera, which results in limited disparities among the viewpoints. We use the built-in Adam optimizer with default parameters and train the entire network over 600 epochs with a batch size of 16. The scale parameter $s$ is initialized as 1 and multiplied by 1.02 for each epoch.

As mentioned earlier, we use the reconstruction loss between the original and reconstructed light fields. However, this is insufficient in some cases. The learned patterns sometimes have significantly different brightness (the brightness of $a^{(n)}$ is given as $\sum_{u,v}a^{(n)}_{u,v}$), which causes too many events spreading over all the pixels. These coding patterns are useless with a real event camera because such a large number of events cannot be recorded correctly due to the limited throughput of the camera. To avoid this problem, we optionally add the second term to the loss function to suppress the number of events in each batch ($N_\mathrm{event}$) below a pre-defined threshold ($\theta$).
\begin{align}
\mathrm{Loss} = \mathrm{MSE}(L, \hat{L}) + \lambda \cdot \max(N_\mathrm{event} - \theta, 0)
\label{eq:loss}
\end{align}
where $\lambda$ is a non-negative weight. The threshold $\theta$ can be computed from the camera's throughput (events per sec.).

\subsection{Hardware}
\label{sec:hardware}

\begin{figure}[t]
\centering
\includegraphics[width=80mm]{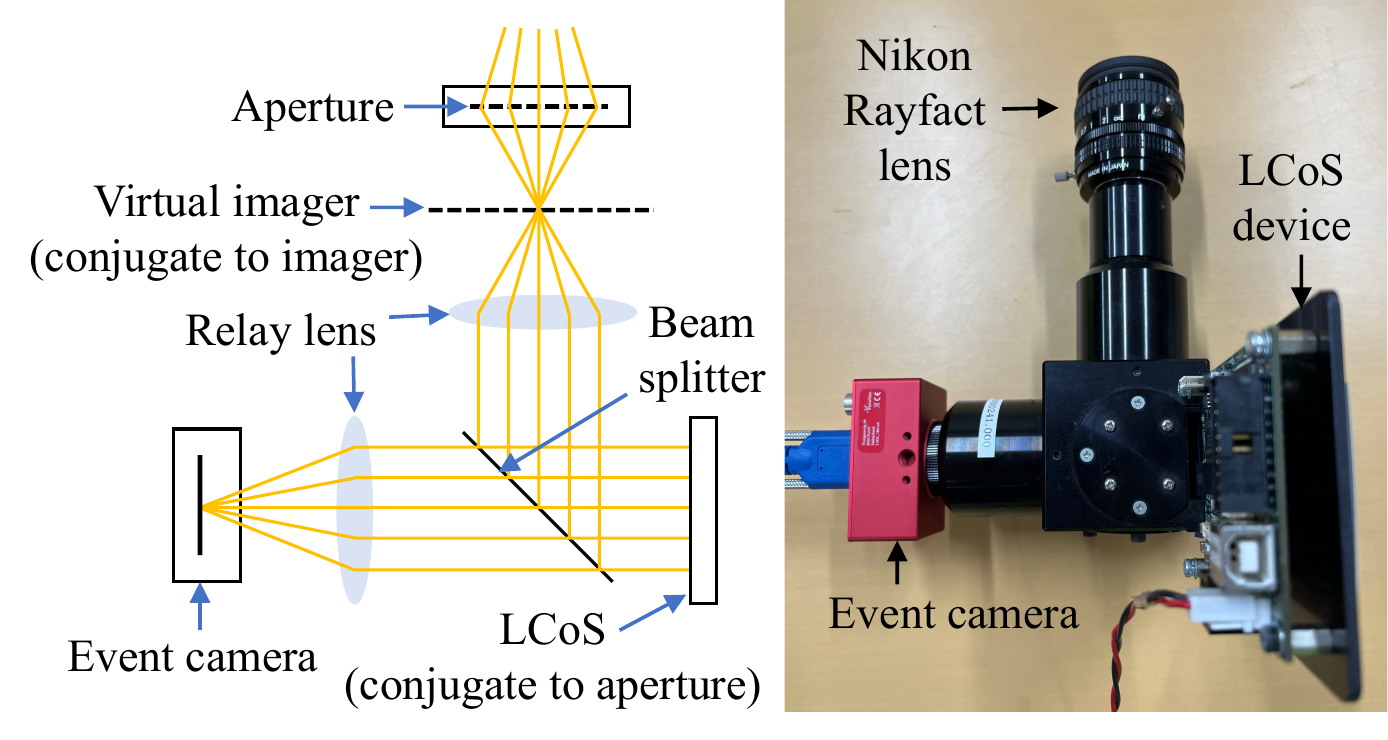}
\vspace{-1mm}
\caption{Hardware setup of our prototype camera.}
\label{fig:hardware}
\includegraphics[width=75mm]{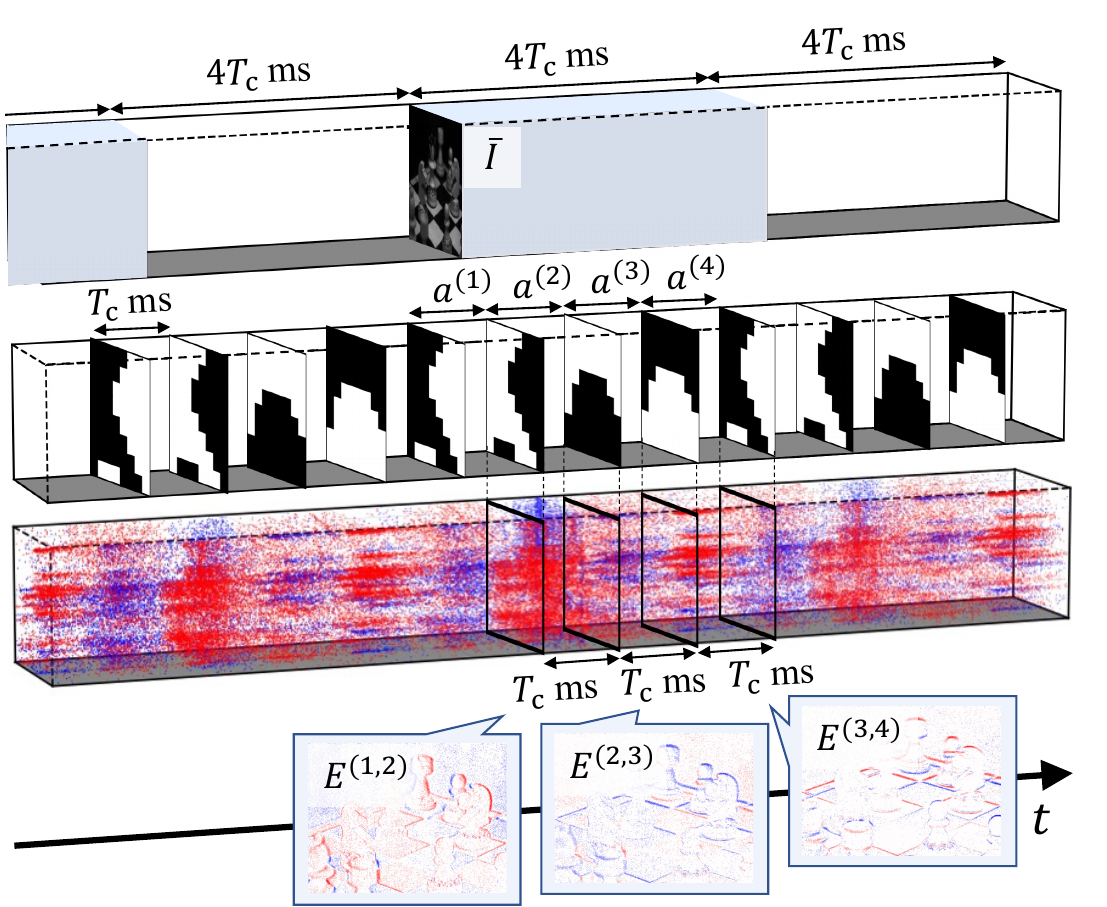}
\caption{Timing chart. From top to bottom, exposure for image frames, aperture-coding patterns, and event stream are shown.}
\label{fig:time-chart}
\end{figure}

Our hardware setup is shown in Fig.~\ref{fig:hardware}. The optical system consists of a Nikon Rayfact lens (25 mm F1.4 SF2514MC), set of relay optics, beam splitter, and LCoS display (Forth Dimension Displays, SXGA-3DM, $1280 \times 1024 $ pixels). We use an iniVation DAVIS 346 monochrome camera that can acquire both events and image frames with $346\times 260$ pixels. 

We use the central portion of the LCoS display ($1024 \times 1024$ pixels) as the effective aperture area, and divide it into $8 \times 8$ regions (each with $128 \times 128$ pixels) to display the coding patterns (each with $8 \times 8$ elements). Four coding patterns, $a^{(1)}$, $a^{(2)}$, $a^{(3)}$, and $a^{(4)}$, are repeatedly displayed, each with a 5.0 msec duration. According to the hardware's document, the total time for each coding pattern ($T_\mathrm{c}$) takes 5.434 msec including the overhead time. The exposure time for an image frame $\bar{I}$ is set to $4T_\mathrm{c}$ ($\simeq$ 22 msec\footnote{This is shorter than those in some previous studies; the exposure time of a coded-aperture camera was set to 40 msec~\cite{Inagaki_2018_ECCV,Sakai_2020_ECCV} and 68 msec~\cite{Mizuno_2022_CVPR}.}) to cover a single cycle of the four coding patterns; See Fig.~\ref{fig:time-chart} for a timing chart. 
Although the aperture coding is not electrically synchronized with the event camera, we can identify the event stacks in the event stream because the cyclic bursts of events correspond to the changes of the coding patterns. 

\section{Experiments}

\subsection{Quantitative Evaluation}

The evaluation was conducted on the test data of the BasicLFSR dataset~\cite{BasicLFSR}, which includes 23 light fields categorized into five groups, all of which were reserved for evaluation. For each light field, we computationally carried out image/event acquisition (AcqNet) and reconstruction (RecNet) processes, and quantitatively evaluated the reconstruction accuracy. Tables \ref{tab:quantitative_comparison} and \ref{tab:quantitative_comparison2} summarize the quantitative scores (PSNR and SSIM, where larger is the better for both) averaged for each group and over all the groups. 

As shown in Table \ref{tab:quantitative_comparison},  we configured four models of our method that are different with respect to the contrast threshold $\tau$. During training, $\tau$ was fixed to 0.075/0.15/0.3 for the fixed-$\tau$-low/mid/high models, respectively, while it was randomly (uniformly) drawn from $[0.075, 0.3]$ for the flexible-$\tau$ model. At test time, $\tau$ was set to 0.075/0.15/0.3. When a fixed-$\tau$ model was trained and tested with an identical $\tau$, a smaller $\tau$ resulted in better reconstruction scores. This is reasonable because with a smaller $\tau$, a larger number of events were observed; thus, more abundant information was obtained from the target scenes. The flexible-$\tau$ model resulted in slightly inferior accuracy to the fixed-$\tau$ models tested with the corresponding $\tau$s. However, the flexible-$\tau$ model was adaptive to various $\tau$s. To make this point clearer, Fig.~\ref{fig:contrast-threshold} shows the reconstruction quality of each model (the average PSNR over all the groups) against a wide range of the test-time $\tau$. The fixed-$\tau$ models were sensitive to the test time $\tau$; each of them had a narrow peak around the $\tau$ with which the model was trained. In contrast, the flexible-$\tau$ model maintained fine reconstruction quality over a wider range of $\tau$, which is important to ensure the compatibility with real event cameras.

\begin{figure}[t]
\centering
\includegraphics[width=.94\linewidth]{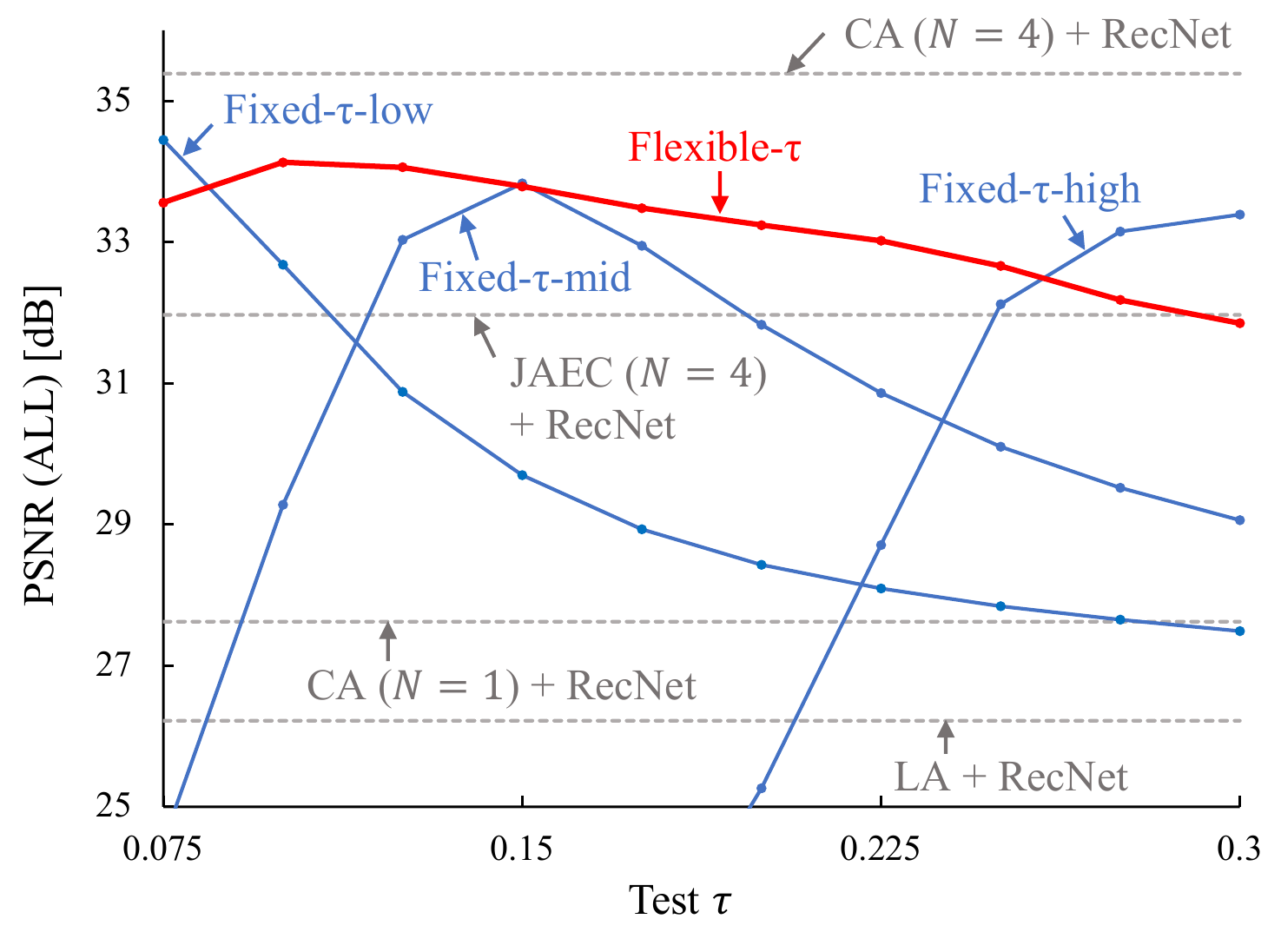}
\vspace{-3mm}
\caption{Reconstruction quality of our method (flexible-$\tau$, fixed-$\tau$-low/mid/high) against test-time $\tau$.}
\label{fig:contrast-threshold}
\end{figure}

\begin{table*}[t]
\centering
    \caption{Quantitative evaluation of \textbf{our method} on BasicLFSR test dataset. Four different models w.r.t. $\tau$ and ablation models are compared. Reported scores are PSNR/SSIM; larger is better for both. $^\dagger$ indicates that model was trained with second loss term.}
    \small
    \label{tab:quantitative_comparison}
    \begin{tabular}{l|c|c||ccccc|c}
      Configuration & train $\tau$ & test $\tau$ 
      & EPFL & HCI (new) & HCI (old) & INRIA & Stanford & ALL \\ \hline
      \hline
      Fixed-$\tau$-low & 0.075 & 0.075 & 34.76/0.9534 & 32.04/0.8522 & 39.53/0.9612 & 36.11/0.9454 & 29.80/0.8973 & 34.45/0.9219 \\
      Fixed-$\tau$-mid & 0.15 & 0.15  & 34.50/0.9503 & 31.22/0.8337 & 38.86/0.9548 & 35.81/0.9437 & 28.78/0.8762 & 33.83/0.9118 \\
      Fixed-$\tau$-high & 0.3 & 0.3  & 34.04/0.9474 & 30.90/0.8287 & 38.41/0.9531 & 35.48/0.9421 & 28.13/0.8684 & 33.39/0.9080 \\ \hline
      & & 0.075 & 34.01/0.9459 & 31.38/0.8434 & 38.41/0.9531 & 35.19/0.9358 & 28.80/0.8783 & 33.56/0.9113 \\
      Flexible-$\tau$ & Random & 0.15 & 34.30/0.9502 & 31.26/0.8424 & 38.95/0.9570 & 35.68/0.9442 & 28.75/0.8806 & 33.79/0.9149 \\
      & & 0.3 & 32.98/0.9348 & 29.60/0.7839 & 35.35/0.9013 & 34.47/0.9358 & 26.87/0.8213 & 31.85/0.8754 \\ \hline
      Image only & -- & --  & 27.35/0.8639 & 25.43/0.7016 & 31.13/0.8462 & 29.08/0.8854 & 21.28/0.6841 & 26.85/0.7962 \\
      Events only$^\dagger$ & Random & 0.15 & 16.29/0.5747 & 12.98/0.4669 & 18.99/0.5835 & 14.92/0.5997 & 10.88/0.4720 & 14.81/0.5394\\
      Events only & Random & 0.15 & 28.40/0.8447 & 27.17/0.7035 & 32.26/0.8300 & 29.43/0.8611 & 24.50/0.7403 & 28.35/0.7963 \\
      
    \end{tabular} \\ \vspace{3mm}

    \caption{Quantitative evaluation of \textbf{other imaging methods} on BasicLFSR test dataset. Second column (``\#'') shows number of acquired images. CA captures one or more images, while full 4-D, JAEC, and LA capture single image.} 
     \label{tab:quantitative_comparison2}
    \begin{tabular}{l|c||ccccc|c}
      Method & \# 
      & EPFL & HCI (new) & HCI (old) & INRIA & Stanford & ALL \\ \hline
      \hline
      CA ($N=4$) + RecNet & 4 & 35.52/0.9556 & 33.06/0.8796 & 40.10/0.9654 & 36.89/0.9471 & 31.39/0.9254 & 35.39/0.9346 \\
      CA ($N=2$) + RecNet & 2 & 34.06/0.9455 & 31.98/0.8599 & 38.82/0.9546 & 35.84/0.9414 & 29.76/0.9037 & 34.09/0.9210 \\
      CA ($N=1$) + RecNet & 1 & 27.78/0.8654 & 26.61/0.7251 & 31.31/0.8352 & 29.40/0.8915 & 22.99/0.7522 & 27.62/0.8139 \\
      \hline
      Full-4D + RecNet & 1 & 32.91/0.9336 & 31.26/0.8371 & 37.90/0.9434 & 34.88/0.9345 & 29.16/0.8895 & 33.22/0.9076\\
      JAEC ($N=4$) + RecNet & 1 & 31.84/0.9195 & 30.05/0.8078 & 36.28/0.9213 & 33.36/0.9256 & 27.80/0.8569 & 31.97/0.8862 \\ 
      JAEC ($N=4$)~\cite{Mizuno_2022_CVPR} & 1 & 30.38/0.9263 & 28.87/0.7732 & 34.96/0.8293 & 33.17/0.9685 & 26.14/0.8473 & 30.70/0.8689 \\ 
      LA + RecNet & 1 & 24.26/0.6843 & 26.17/0.6714 & 30.81/0.7978 & 25.85/0.7628 & 24.03/0.6926 & 26.22/0.7218 \\
      LA (naive) & 1 & 22.25/0.5820 & 24.75/0.6050 & 28.65/0.7157 & 23.71/0.6829 & 22.42/0.5934 & 24.36/0.6358 \\

    \end{tabular}

\end{table*}

Table \ref{tab:quantitative_comparison} also includes ablation models with which the measurement was limited to either the image frame (image-only) or the event stacks (event-only); for each ablation model, the same algorithm pipeline as ours was trained from scratch. The second term of the loss function (Eq.~(\ref{eq:loss}), with $\lambda=0.00001$ and $\theta=$ 131,130) was enabled only for the event-only model, which is indicated as ``$^\dagger$''. Without this term, the learned coding patterns of the event-only model had significant brightness differences and caused too many events. The poor reconstruction quality obtained with the ablation models indicates that both the image frame and events are necessary for accurate light-field reconstruction.

\begin{figure*}[t]
\centering
\begin{tabular}{cccc}
\includegraphics[width=40mm]{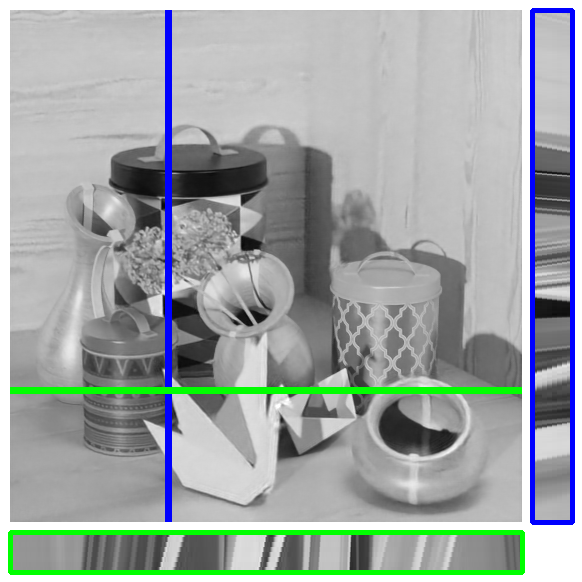} &
\includegraphics[width=40mm]{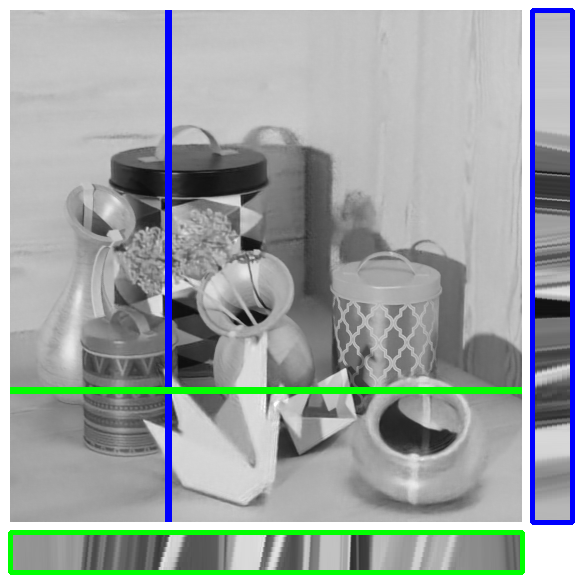} &
\includegraphics[width=40mm]{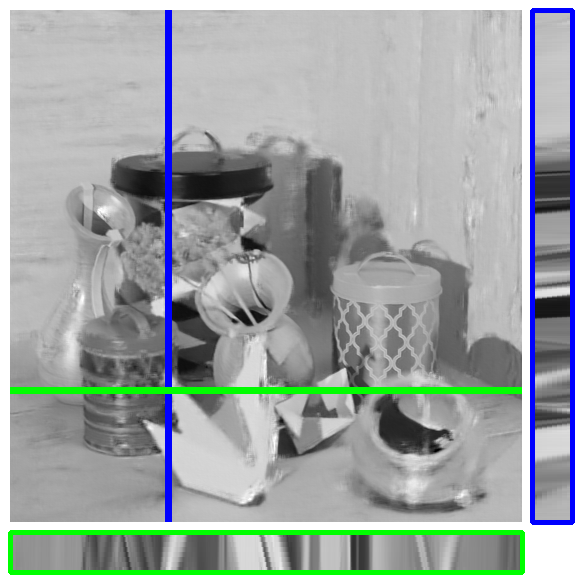} &
\includegraphics[width=40mm]{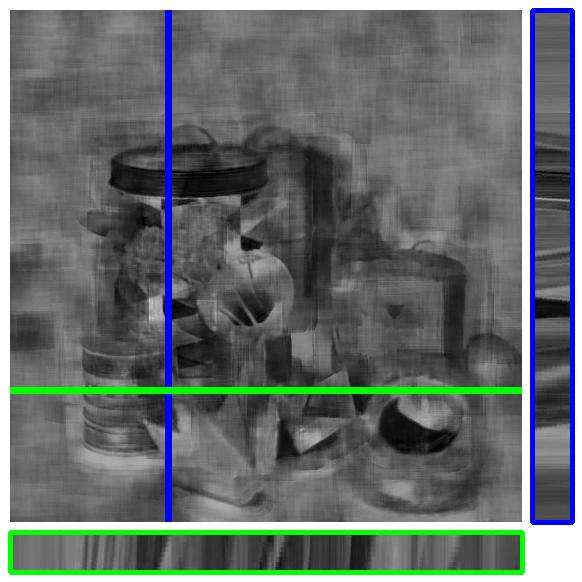} \\
\includegraphics[width=20mm]{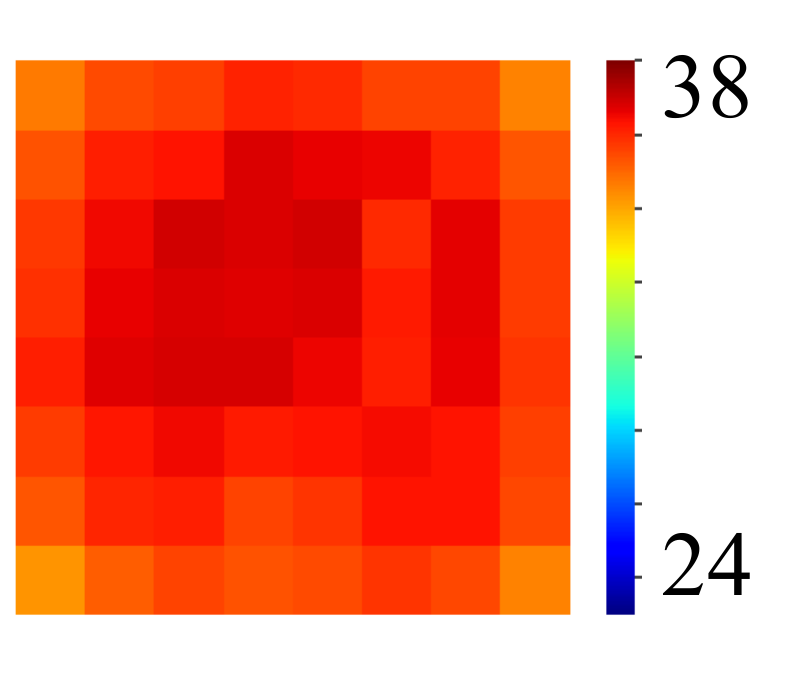} &
\includegraphics[width=20mm]{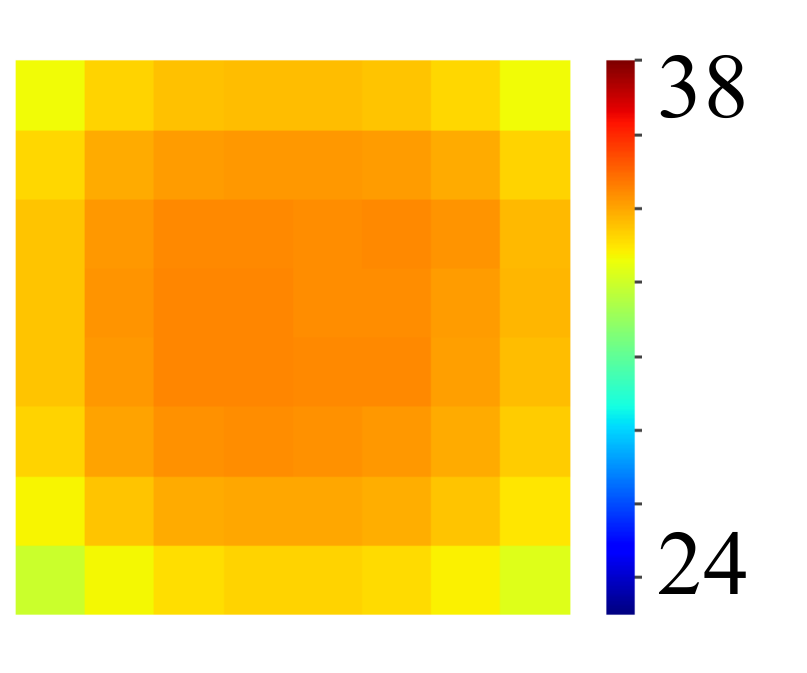} &
\includegraphics[width=20mm]{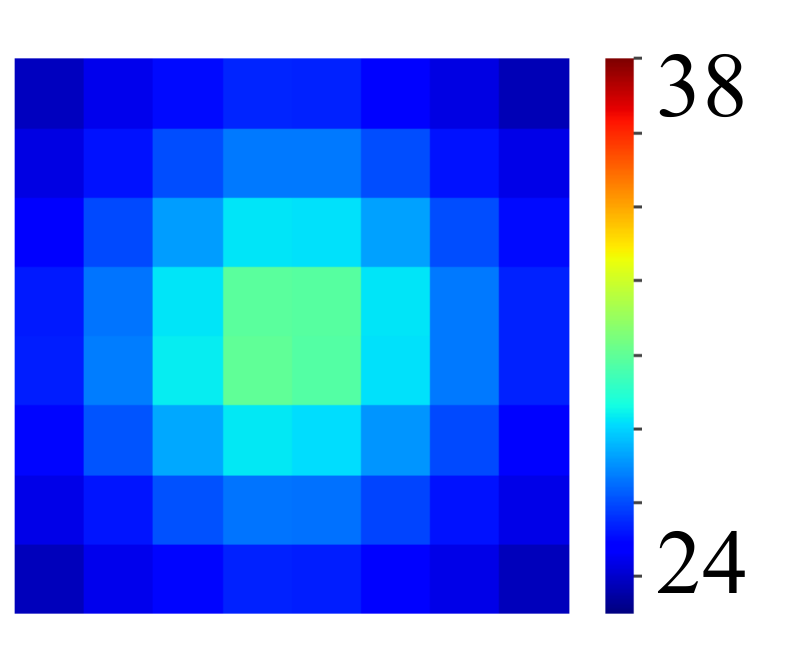} &
\includegraphics[width=20mm]{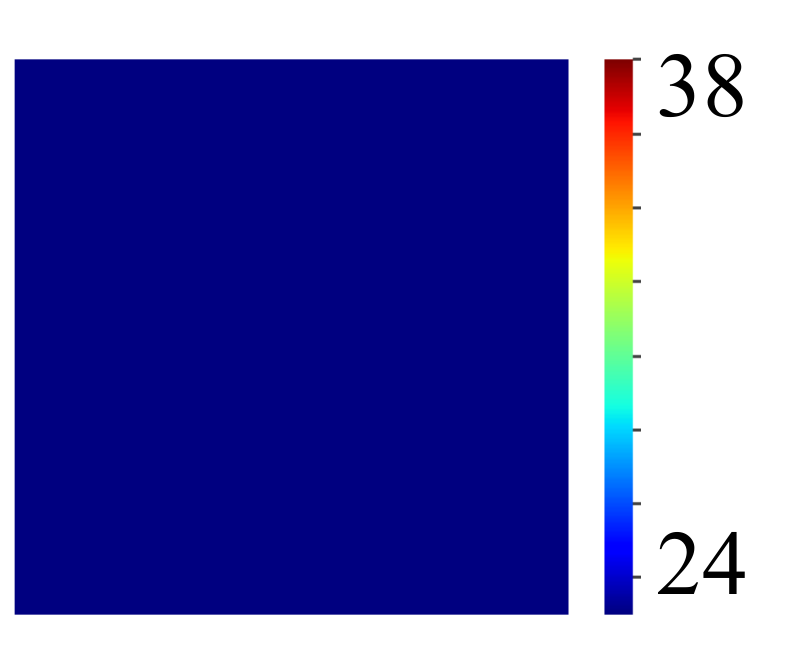} \\
{\small Coded aperture ($N=4$)} & {\small Ours (flex., test $\tau=0.15$)} & {\small Ours (image only)} & {\small Ours (event only$^\dagger$)} \\
\end{tabular}

\caption{Visual result tested on \textit{Origami} light field. Top: reconstructed top-left views with epipolar plane images along blue/green lines. Bottom: view-by-view reconstruction quality in PSNR. }
\label{fig:visual-result}
\vspace{3mm}
\begin{minipage}{0.28\linewidth}
\centering
\includegraphics[width=11.2mm]{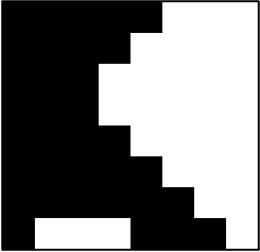}
\includegraphics[width=11.2mm]{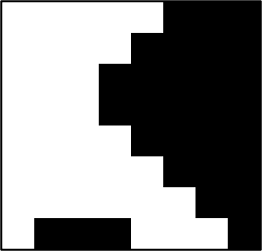}
\includegraphics[width=11.2mm]{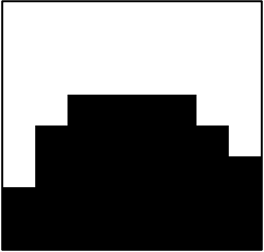}
\includegraphics[width=11.2mm]{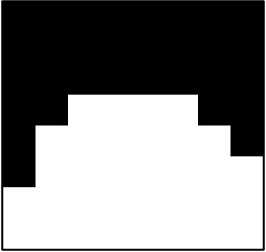}\\
\includegraphics[width=47.5mm]{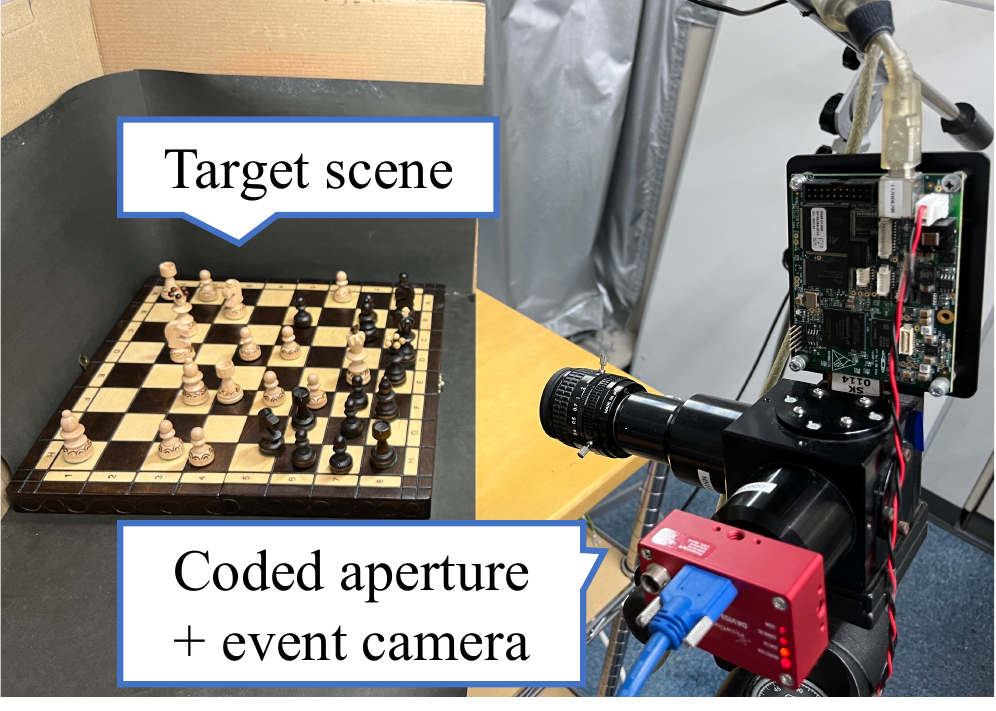}
\end{minipage}
\begin{minipage}{0.35\linewidth}
\centering
\includegraphics[width=30mm]{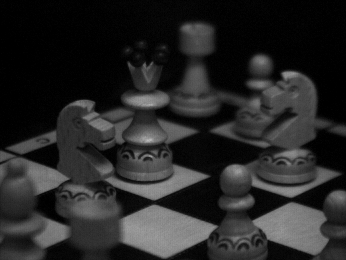}
\includegraphics[width=30mm]{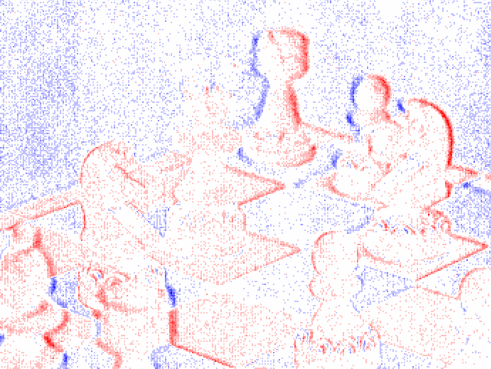} 
\includegraphics[width=30mm]{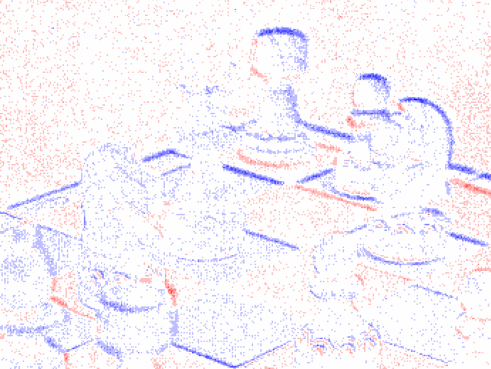}
\includegraphics[width=30mm]{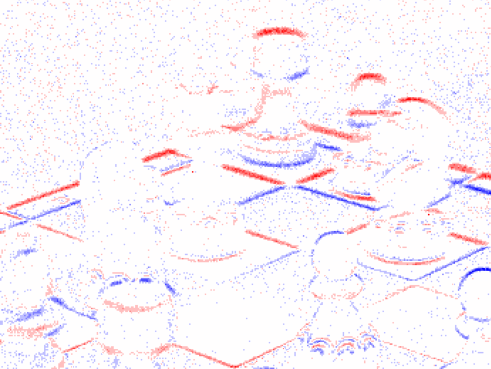}
\end{minipage}
\begin{minipage}{0.05\linewidth}
\includegraphics[width=5mm]{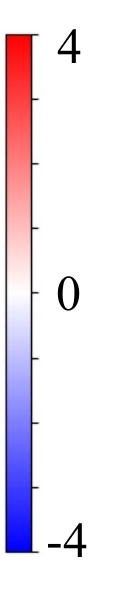}
\end{minipage}
\hspace{-6mm}
\begin{minipage}{0.3\linewidth}
\includegraphics[width=55mm]{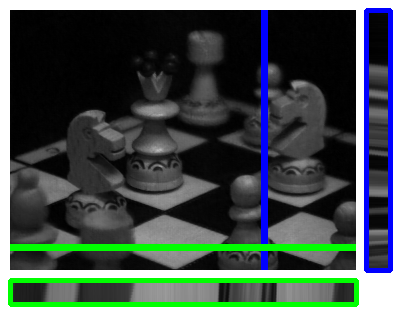}
\end{minipage}
\caption{Experiment with our prototype camera. Coding patterns and experimental setup (left), image frame and event stacks obtained from camera (center), and reconstructed light field view (at top-left viewpoint) with epipolar plane images along blue/green lines (right).}
\label{fig:real-exp}
\end{figure*}

As summarized in Table \ref{tab:quantitative_comparison2}, we also tested other imaging methods for comparison. As the baseline of our method, \textbf{coded-aperture imaging (CA)}~\cite{liang2008programmable,nagahara2010programmable,Inagaki_2018_ECCV,Sakai_2020_ECCV,Guo2022TPAMI} takes several ($N$) images following Eq.~(\ref{eq:coded-aperture}), which requires a longer measurement time than a single exposure. \textbf{Joint aperture-exposure coding (JAEC)}~\cite{Tateishi_2021_ICIP,vargas_2021_ICCV,Mizuno_2022_CVPR,Tateishi_2022_IEICE} captures a single coded image following Eq.~(\ref{eq:joint-coding}) with $N=4$.  \textbf{Full-4D}~\cite{Tateishi_2022_IEICE} is an idealized hypothetical imaging model without physical hardware implementations, which enables arbitrary 4-D coding while capturing a single image:
\begin{eqnarray}
I_{x,y} = \sum_{u,v} m_{x,y,u, v}L_{x,y,u,v}
\end{eqnarray}
where $m_{x,y,u,v} \in [0,1]$. With these three methods, we appended the same amount of noise as our method to the observed images ($\sigma = 0.005$ w.r.t. the intensity range $[0,1]$). As indicated by ``+RecNet'', each method was combined with a light-field reconstruction network that had the same architecture as RecNet.\footnote{With CA, $N$ observed images were stacked along the channel dimension and fed to RecNet. With JAEC and full4D, we lifted the observed image to a sparse tensor with 64 channels before feeding to RecNet, as was done in a previous study~\cite{Tateishi_2022_IEICE}.} The network was trained for each imaging method from scratch; the coding patterns and weights in RecNet were jointly optimized on the same dataset as ours.\footnote{With JAEC, we fixed the coding patterns for the imaging plane, $p^{1},\dots,p^{4}$, to those provided by Mizuno et al.'~\cite{Mizuno_2022_CVPR} to ensure the compatibility with the hardware constraints.} We also used the software of JAEC provided by Mizuno et al.~\cite{Mizuno_2022_CVPR}, which had 12 times the parameters of our RecNet, and was retrained on the same dataset as ours. Finally, to simulate \textbf{lens-array based imaging (LA)}~\cite{adelson1992single,arai1998gradient,ng2005light,ng2006digital}, which can take a light field in a single shot, we down-sampled each view of a light field into the 1/8$\times$1/8 spatial resolution, and up-sampled it into the original resolution using bicubic interpolation (``naive''). We also used RecNet to enhance the quality of the up-sampled light field (``+RecNet'').\footnote{The up-sampled 64 views were stacked along the channel dimension and fed to RecNet, and RecNet was trained on the same dataset as ours.}

CA ($N=4$) can be regarded as the upper-bound reference for our method since our method can obtain a set of data that is \textit{quasi-equivalent}  to the four coded-aperture images (See \ref{sec:theory}). Aligned with this theory, the quantitative scores of our method (fixed-$\tau$-low) were close to those of CA ($N=4$). Our method with a moderate configuration (flexible-$\tau$, test $\tau = 0.15$) still performed comparably to CA ($N=2$) and outperformed CA ($N=1$), full-4D, JAEC, and LA. Our method (flexible-$\tau$) was also stable over a wide range of $\tau$. As shown in Fig.~\ref{fig:contrast-threshold}. our method (flexible-$\tau$) consistently ($\tau \in [0.075, 0.275]$) outperformed the other imaging methods that can complete the measurement in a single exposure (CA~($N=1$), JAEC, and LA).

Several visual results are presented in Fig.~\ref{fig:visual-result}. Our method obtained a visually-convincing result comparable to that of the reference (CA with $N=4$). The image-only model reconstructed the overall appearance but lost some details and the consistency among the viewpoints. The result obtained with the events-only$^\dagger$ model seems somewhat consistent among the viewpoints but lacking correct intensity. Please refer to the supplementary video for more results with better visualization.

\subsection{Real-World Experiment}

To capture real 3-D scenes, we used our prototype camera and flexible-$\tau$ model. The aperture-coding patterns were set to the corresponding parameters in the learned AcqNet model. Figure~\ref{fig:real-exp} shows the coding patterns and experimental setup (left), an image frame and event stacks obtained from the camera (center), and the light field reconstructed from them (right). Our method performed well with real 3-D scenes; the parallax information was successfully embedded into the events, and a light field was reconstructed with visually convincing quality. Please refer to the supplementary video for more results with better visualization.

\section{Conclusion}

We proposed a new light-field acquisition method that takes advantage of coded-aperture imaging and an event camera. Although the measurement on the camera can be completed in a single exposure, our method is theoretically quasi-equivalent to the baseline coded-aperture imaging method. We carefully designed our algorithm pipeline to be end-to-end trainable and compatible with real camera hardware, which enabled both accurate light-field reconstruction and successful development of our hardware prototype. 

For our future work, we will explore the design space with respect to, e.g., the number of coding patterns during an exposure, while considering the hardware restrictions. We will also extend our method to moving scenes (light field videos).
Seeking other possible means for hardware implementation to achieve RGB color reconstruction and higher spatial resolutions is also an important avenue. 
Finally, our method would be extended to a more event-centered direction, where event streams over continuous time (instead of event stacks) could be fully utilized and light fields could be reconstructed from events alone.

\noindent\textbf{Acknowledgement:} This work was supported by JSPS Grant-in-Aid for Scientific Research (B) 22H03611 and NICT contract research JPJ012368C06801.

\small
\bibliographystyle{ieeenat_fullname}
\bibliography{refs}

\begin{thebibliography}{57}
\providecommand{\natexlab}[1]{#1}
\providecommand{\url}[1]{\texttt{#1}}
\expandafter\ifx\csname urlstyle\endcsname\relax
  \providecommand{\doi}[1]{doi: #1}\else
  \providecommand{\doi}{doi: \begingroup \urlstyle{rm}\Url}\fi

\bibitem[Adelson and Wang(1992)]{adelson1992single}
Edward~H Adelson and John~YA Wang.
\newblock Single lens stereo with a plenoptic camera.
\newblock \emph{IEEE transactions on pattern analysis and machine
  intelligence}, 14\penalty0 (2):\penalty0 99--106, 1992.

\bibitem[Arai et~al.(1998)Arai, Okano, Hoshino, and Yuyama]{arai1998gradient}
Jun Arai, Fumio Okano, Haruo Hoshino, and Ichiro Yuyama.
\newblock Gradient-index lens-array method based on real-time integral
  photography for three-dimensional images.
\newblock \emph{Applied optics}, 37\penalty0 (11):\penalty0 2034--2045, 1998.

\bibitem[Br\"{a}ndli et~al.(2014)Br\"{a}ndli, Berner, Yang, Liu, and
  Delbruck]{Brandli_2014_IEEE}
Christian Br\"{a}ndli, Raphael Berner, Minhao Yang, Shih-Chii Liu, and Tobi
  Delbruck.
\newblock A 240 × 180 130 {dB} 3 $\mu s$ latency global shutter spatiotemporal
  vision sensor.
\newblock \emph{IEEE Journal of Solid-State Circuits}, 49:\penalty0 2333--2341,
  2014.

\bibitem[Broxton et~al.(2020)Broxton, Flynn, Overbeck, Erickson, Hedman,
  DuVall, Dourgarian, Busch, Whalen, and Debevec]{broxton2020immersive}
Michael Broxton, John Flynn, Ryan Overbeck, Daniel Erickson, Peter Hedman,
  Matthew DuVall, Jason Dourgarian, Jay Busch, Matt Whalen, and Paul Debevec.
\newblock Immersive light field video with a layered mesh representation.
\newblock In \emph{{ACM} Transactions on Graphics (Proc. SIGGRAPH)}, 2020.

\bibitem[Chan et~al.(2021)Chan, Monteiro, Kellnhofer, Wu, and
  Wetzstein]{Chan_2021_CVPR}
Eric~R. Chan, Marco Monteiro, Petr Kellnhofer, Jiajun Wu, and Gordon Wetzstein.
\newblock {Pi-GAN}: Periodic implicit generative adversarial networks for
  {3D-Aware} image synthesis.
\newblock In \emph{Proceedings of the IEEE/CVF Conference on Computer Vision
  and Pattern Recognition (CVPR)}, 2021.

\bibitem[Chan et~al.(2023)Chan, Nagano, Chan, Bergman, Park, Levy, Aittala,
  De~Mello, Karras, and Wetzstein]{Chan_2023_ICCV}
Eric~R. Chan, Koki Nagano, Matthew~A. Chan, Alexander~W. Bergman, Jeong~Joon
  Park, Axel Levy, Miika Aittala, Shalini De~Mello, Tero Karras, and Gordon
  Wetzstein.
\newblock Generative novel view synthesis with {3D}-aware diffusion models.
\newblock In \emph{Proceedings of the IEEE/CVF International Conference on
  Computer Vision (ICCV)}, pages 4217--4229, 2023.

\bibitem[Chen et~al.(2020)Chen, Ruan, and Lam]{Chen2020LFGAN}
Bin Chen, Lingyan Ruan, and Miu-Ling Lam.
\newblock {LFGAN}: {4D} light field synthesis from a single {RGB} image.
\newblock \emph{ACM Trans. Multimedia Comput. Commun. Appl.}, 16\penalty0 (1),
  2020.

\bibitem[Fujii et~al.(2006)Fujii, Mori, Takeda, Mase, Tanimoto, and
  Suenaga]{fujii2006multipoint}
Toshiaki Fujii, Kensaku Mori, Kazuya Takeda, Kenji Mase, Masayuki Tanimoto, and
  Yasuhito Suenaga.
\newblock Multipoint measuring system for video and sound - 100-camera and
  microphone system.
\newblock In \emph{IEEE International Conference on Multimedia and Expo}, pages
  437--440, 2006.

\bibitem[Gallego et~al.(2022)Gallego, Delbr{\"u}ck, Orchard, Bartolozzi, Taba,
  Censi, Leutenegger, Davison, Conradt, Daniilidis, and
  Scaramuzza]{eventcamerasurvey2022}
Guillermo Gallego, Tobi Delbr{\"u}ck, Garrick Orchard, Chiara Bartolozzi, Brian
  Taba, Andrea Censi, Stefan Leutenegger, Andrew~J. Davison, J{\"u}rg Conradt,
  Kostas Daniilidis, and Davide Scaramuzza.
\newblock Event-based vision: A survey.
\newblock \emph{IEEE Transactions on Pattern Analysis and Machine
  Intelligence}, 44\penalty0 (1):\penalty0 154--180, 2022.

\bibitem[Guo et~al.(2022)Guo, Hou, Jin, Chen, and Chau]{Guo2022TPAMI}
Mantang Guo, Junhui Hou, Jing Jin, Jie Chen, and Lap-Pui Chau.
\newblock Deep spatial-angular regularization for light field imaging,
  denoising, and super-resolution.
\newblock \emph{IEEE Transactions on Pattern Analysis and Machine
  Intelligence}, 44\penalty0 (10):\penalty0 6094--6110, 2022.

\bibitem[Honauer et~al.(2016)Honauer, Johannsen, Kondermann, and
  Goldluecke]{honauer2017dataset}
Katrin Honauer, Ole Johannsen, Daniel Kondermann, and Bastian Goldluecke.
\newblock A dataset and evaluation methodology for depth estimation on {4D}
  light fields.
\newblock In \emph{Asian Conference on Computer Vision}, 2016.

\bibitem[Huang et~al.(2015)Huang, Chen, and Wetzstein]{huang2015light}
Fu-Chung Huang, Kevin Chen, and Gordon Wetzstein.
\newblock The light field stereoscope: immersive computer graphics via factored
  near-eye light field displays with focus cues.
\newblock \emph{ACM Transactions on Graphics}, 34\penalty0 (4):\penalty0 60,
  2015.

\bibitem[Iliadis et~al.(2016)Iliadis, Spinoulas, and
  Katsaggelos]{Iliadis_2016_mask}
Michael Iliadis, Leonidas Spinoulas, and Aggelos~K. Katsaggelos.
\newblock Deepbinarymask: Learning a binary mask for video compressive sensing,
  2016.

\bibitem[Inagaki et~al.(2018)Inagaki, Kobayashi, Takahashi, Fujii, and
  Nagahara]{Inagaki_2018_ECCV}
Yasutaka Inagaki, Yuto Kobayashi, Keita Takahashi, Toshiaki Fujii, and Hajime
  Nagahara.
\newblock Learning to capture light fields through a coded aperture camera.
\newblock In \emph{European Conference on Computer Vision}, pages 418--434,
  2018.

\bibitem[Kalantari et~al.(2016)Kalantari, Wang, and Ramamoorthi]{Kalantari2016}
Nima~Khademi Kalantari, Ting-Chun Wang, and Ravi Ramamoorthi.
\newblock Learning-based view synthesis for light field cameras.
\newblock \emph{ACM Transactions on Graphics}, 35\penalty0 (6), 2016.

\bibitem[Khan et~al.(2021)Khan, Kim, and Tompkin]{khan2021edgeaware}
Numair Khan, Min~H. Kim, and James Tompkin.
\newblock Edge-aware bidirectional diffusion for dense depth estimation from
  light fields.
\newblock In \emph{British Machine Vision Conference (BMVC)}, 2021.

\bibitem[Kim et~al.(2016)Kim, Leutenegger, and Davison]{Kim2016Real}
H Kim, S Leutenegger, and AJ Davison.
\newblock Real-time {3D} reconstruction and 6-{DoF} tracking with an event
  camera.
\newblock In \emph{European Conference on Computer Vision (ECCV)}, pages
  349--364, 2016.

\bibitem[Lee et~al.(2016)Lee, Jang, Moon, Cho, and Lee]{lee2016additive}
Seungjae Lee, Changwon Jang, Seokil Moon, Jaebum Cho, and Byoungho Lee.
\newblock Additive light field displays: realization of augmented reality with
  holographic optical elements.
\newblock \emph{ACM Transactions on Graphics}, 35\penalty0 (4):\penalty0 1--13,
  2016.

\bibitem[Li et~al.(2021)Li, Feng, She, Ding, Wang, and Lee]{mine2021ICCV}
Jiaxin Li, Zijian Feng, Qi She, Henghui Ding, Changhu Wang, and Gim~Hee Lee.
\newblock {MINE}: Towards continuous depth {MPI} with {NeRF} for novel view
  synthesis.
\newblock In \emph{International Conference on Computer Vision}, 2021.

\bibitem[Li and Khademi~Kalantari(2020)]{VariableMPI2020ACM}
Qinbo Li and Nima Khademi~Kalantari.
\newblock Synthesizing light field from a single image with variable {MPI} and
  two network fusion.
\newblock \emph{{ACM} Transactions on Graphics}, 2020.

\bibitem[Li et~al.(2020)Li, Qi, Gulve, Wei, Genov, Kutulakos, and
  Heidrich]{Li_2020_ICCP}
Yuqi Li, Miao Qi, Rahul Gulve, Mian Wei, Roman Genov, Kiriakos~N. Kutulakos,
  and Wolfgang Heidrich.
\newblock End-to-end video compressive sensing using anderson-accelerated
  unrolled networks.
\newblock In \emph{International Conference on Computational Photography},
  pages 137--148, 2020.

\bibitem[Liang et~al.(2008)Liang, Lin, Wong, Liu, and
  Chen]{liang2008programmable}
Chia-Kai Liang, Tai-Hsu Lin, Bing-Yi Wong, Chi Liu, and Homer~H Chen.
\newblock Programmable aperture photography: multiplexed light field
  acquisition.
\newblock \emph{ACM Transactions on Graphics}, 27\penalty0 (3):\penalty0 1--10,
  2008.

\bibitem[Liang(2021)]{BasicLFSR}
Zhengyu Liang.
\newblock Basic{LFSR} (open source light field toolbox for super-resolution).
\newblock \url{https://github.com/ZhengyuLiang24/BasicLFSR}, 2021.

\bibitem[Ma et~al.(2023)Ma, Paudel, Chhatkuli, and Van~Gool]{Ma_2023_ICCV}
Qi Ma, Danda~Pani Paudel, Ajad Chhatkuli, and Luc Van~Gool.
\newblock Deformable neural radiance fields using {RGB} and event cameras.
\newblock In \emph{Proceedings of the IEEE/CVF International Conference on
  Computer Vision (ICCV)}, pages 3590--3600, 2023.

\bibitem[Maeno et~al.(2013)Maeno, Nagahara, Shimada, and Taniguchi]{Maeno2013}
Kazuki Maeno, Hajime Nagahara, Atsushi Shimada, and Rin-Ichiro Taniguchi.
\newblock Light field distortion feature for transparent object recognition.
\newblock In \emph{IEEE Conference on Computer Vision and Pattern Recognition},
  pages 2786--2793, 2013.

\bibitem[Marwah et~al.(2013)Marwah, Wetzstein, Bando, and
  Raskar]{marwah2013compressive}
Kshitij Marwah, Gordon Wetzstein, Yosuke Bando, and Ramesh Raskar.
\newblock Compressive light field photography using overcomplete dictionaries
  and optimized projections.
\newblock \emph{ACM Transactions on Graphics}, 32\penalty0 (4):\penalty0 1--12,
  2013.

\bibitem[Mildenhall et~al.(2019)Mildenhall, Srinivasan, Ortiz-Cayon, Kalantari,
  Ramamoorthi, Ng, and Kar]{mildenhall2019llff}
Ben Mildenhall, Pratul~P. Srinivasan, Rodrigo Ortiz-Cayon, Nima~Khademi
  Kalantari, Ravi Ramamoorthi, Ren Ng, and Abhishek Kar.
\newblock Local light field fusion: Practical view synthesis with prescriptive
  sampling guidelines.
\newblock \emph{ACM Transactions on Graphics}, 38:\penalty0 1--14, 2019.

\bibitem[Mizuno et~al.(2022)Mizuno, Takahashi, Yoshida, Tsutake, Fujii, and
  Nagahara]{Mizuno_2022_CVPR}
Ryoya Mizuno, Keita Takahashi, Michitaka Yoshida, Chihiro Tsutake, Toshiaki
  Fujii, and Hajime Nagahara.
\newblock Acquiring a dynamic light field through a single-shot coded image.
\newblock In \emph{Proceedings of the IEEE/CVF Conference on Computer Vision
  and Pattern Recognition (CVPR)}, 2022.

\bibitem[Nabati et~al.(2018)Nabati, Mendlovic, and Giryes]{nabati2018colorLF}
Ofir Nabati, David Mendlovic, and Raja Giryes.
\newblock Fast and accurate reconstruction of compressed color light field.
\newblock In \emph{International Conference on Computational Photography},
  pages 1--11, 2018.

\bibitem[Nagahara et~al.(2010)Nagahara, Zhou, Watanabe, Ishiguro, and
  Nayar]{nagahara2010programmable}
Hajime Nagahara, Changyin Zhou, Takuya Watanabe, Hiroshi Ishiguro, and Shree~K
  Nayar.
\newblock Programmable aperture camera using {LCoS}.
\newblock In \emph{European Conference on Computer Vision}, pages 337--350,
  2010.

\bibitem[Ng(2006)]{ng2006digital}
Ren Ng.
\newblock \emph{Digital light field photography}.
\newblock PhD thesis, Stanford University, 2006.

\bibitem[Ng et~al.(2005)Ng, Levoy, Br{\'e}dif, Duval, Horowitz, and
  Hanrahan]{ng2005light}
Ren Ng, Marc Levoy, Mathieu Br{\'e}dif, Gene Duval, Mark Horowitz, and Pat
  Hanrahan.
\newblock Light field photography with a hand-held plenoptic camera.
\newblock \emph{Computer Science Technical Report CSTR}, 2\penalty0
  (11):\penalty0 1--11, 2005.

\bibitem[Nie et~al.(2018)Nie, Gu, Zheng, Lam, Ono, and Sato]{Nie_2018_CVPR}
Shijie Nie, Lin Gu, Yinqiang Zheng, Antony Lam, Nobutaka Ono, and Imari Sato.
\newblock Deeply learned filter response functions for hyperspectral
  reconstruction.
\newblock In \emph{IEEE/CVF Conference on Computer Vision and Pattern
  Recognition}, pages 4767--4776, 2018.

\bibitem[Niklaus et~al.(2019)Niklaus, Mai, Yang, and Liu]{Niklaus_TOG_2019}
Simon Niklaus, Long Mai, Jimei Yang, and Feng Liu.
\newblock {3D} ken burns effect from a single image.
\newblock \emph{ACM Transactions on Graphics}, 38\penalty0 (6):\penalty0
  184:1--184:15, 2019.

\bibitem[Rebecq et~al.(2018{\natexlab{a}})Rebecq, Gallego, Muggler, and
  Scaramuzza]{Rebecq2018EMVS}
Henri Rebecq, Guillermo Gallego, Elias Muggler, and Davide Scaramuzza.
\newblock {EMVS}: Event-based multi-view stereo -- {3D} reconstruction with an
  event camera in real-time.
\newblock \emph{International Journal of Computer Vision}, 126\penalty0
  (12):\penalty0 1394--1414, 2018{\natexlab{a}}.

\bibitem[Rebecq et~al.(2018{\natexlab{b}})Rebecq, Gehrig, and
  Scaramuzza]{Rebecq18corl}
Henri Rebecq, Daniel Gehrig, and Davide Scaramuzza.
\newblock {ESIM}: an open event camera simulator.
\newblock \emph{Conf. on Robotics Learning (CoRL)}, 2018{\natexlab{b}}.

\bibitem[Rudnev et~al.(2023)Rudnev, Elgharib, Theobalt, and
  Golyanik]{rudnev2023eventnerf}
Viktor Rudnev, Mohamed Elgharib, Christian Theobalt, and Vladislav Golyanik.
\newblock {EventNeRF}: Neural radiance fields from a single colour event
  camera.
\newblock In \emph{Computer Vision and Pattern Recognition (CVPR)}, 2023.

\bibitem[Sakai et~al.(2020)Sakai, Takahashi, Fujii, and
  Nagahara]{Sakai_2020_ECCV}
Kohei Sakai, Keita Takahashi, Toshiaki Fujii, and Hajime Nagahara.
\newblock Acquiring dynamic light fields through coded aperture camera.
\newblock In \emph{European Conference on Computer Vision}, pages 368--385,
  2020.

\bibitem[Shin et~al.(2018)Shin, Jeon, Yoon, Kweon, and Kim]{shin18epinet}
Changha Shin, Hae-Gon Jeon, Youngjin Yoon, In~So Kweon, and Seon~Joo Kim.
\newblock {EPINET}: A fully-convolutional neural network using epipolar
  geometry for depth from light field images.
\newblock In \emph{IEEE Conference on Computer Vision and Pattern Recognition},
  pages 4748--4757, 2018.

\bibitem[Taguchi et~al.(2009)Taguchi, Koike, Takahashi, and
  Naemura]{Taguchi2009}
Yuichi Taguchi, Takafumi Koike, Keita Takahashi, and Takeshi Naemura.
\newblock Trans{CAIP}: A live 3{D} {TV} system using a camera array and an
  integral photography display with interactive control of viewing parameters.
\newblock \emph{IEEE Transactions on Visualization and Computer Graphics},
  15\penalty0 (5):\penalty0 841--852, 2009.

\bibitem[Tateishi et~al.(2021)Tateishi, Sakai, Tsutake, Takahashi, and
  Fujii]{Tateishi_2021_ICIP}
Kohei Tateishi, Kohei Sakai, Chihiro Tsutake, Keita Takahashi, and Toshiaki
  Fujii.
\newblock Factor modulation for single-shot light-field acquisition.
\newblock In \emph{IEEE International Conference on Image Processing}, pages
  3253--3257, 2021.

\bibitem[Tateishi et~al.(2022)Tateishi, Tsutake, Takahashi, and
  Fujii]{Tateishi_2022_IEICE}
Kohei Tateishi, Chihiro Tsutake, Keita Takahashi, and Toshiaki Fujii.
\newblock Time-multiplexed coded aperture and coded focal stack -comparative
  study on snapshot compressive light field imaging.
\newblock \emph{IEICE Transactions on Information and Systems}, E105.D\penalty0
  (10):\penalty0 1679--1690, 2022.

\bibitem[Trevithick and Yang(2021)]{Trevithick_2021_ICCV}
Alex Trevithick and Bo Yang.
\newblock {GRF}: Learning a general radiance field for {3D} representation and
  rendering.
\newblock In \emph{International Conference on Computer Vision}, 2021.

\bibitem[Tucker and Snavely(2020)]{single_view_mpi}
Richard Tucker and Noah Snavely.
\newblock Single-view view synthesis with multiplane images.
\newblock In \emph{IEEE Conference on Computer Vision and Pattern Recognition},
  2020.

\bibitem[Vadathya et~al.(2019)Vadathya, Girish, and Mitra]{vadathya2019}
Anil~Kumar Vadathya, Sharath Girish, and Kaushik Mitra.
\newblock A unified learning based framework for light field reconstruction
  from coded projections.
\newblock \emph{IEEE Transactions on Computational Imaging}, 6:\penalty0
  304--316, 2019.

\bibitem[Vargas et~al.(2021)Vargas, Martel, Wetzstein, and
  Arguello]{vargas_2021_ICCV}
Edwin Vargas, Julien N.~P. Martel, Gordon Wetzstein, and Henry Arguello.
\newblock Time-multiplexed coded aperture imaging: Learned coded aperture and
  pixel exposures for compressive imaging systems.
\newblock In \emph{International Conference on Computer Vision}, 2021.

\bibitem[Veeraraghavan et~al.(2007)Veeraraghavan, Raskar, Agrawal, Mohan, and
  Tumblin]{veeraraghavan2007dappled}
Ashok Veeraraghavan, Ramesh Raskar, Amit Agrawal, Ankit Mohan, and Jack
  Tumblin.
\newblock Dappled photography: Mask enhanced cameras for heterodyned light
  fields and coded aperture refocusing.
\newblock \emph{ACM Transactions on Graphics}, 26\penalty0 (3):\penalty0 69,
  2007.

\bibitem[Wang et~al.(2023)Wang, Wang, and Yuan]{Wang_2023_ICCV}
Ping Wang, Lishun Wang, and Xin Yuan.
\newblock Deep optics for video snapshot compressive imaging.
\newblock In \emph{Proceedings of the IEEE/CVF International Conference on
  Computer Vision (ICCV)}, pages 10646--10656, 2023.

\bibitem[Wang et~al.(2016)Wang, Zhu, Hiroaki, Chandraker, Efros, and
  Ramamoorthi]{wang20164d}
Ting-Chun Wang, Jun-Yan Zhu, Ebi Hiroaki, Manmohan Chandraker, Alexei~A Efros,
  and Ravi Ramamoorthi.
\newblock A {4D} light-field dataset and {CNN} architectures for material
  recognition.
\newblock In \emph{European Conference on Computer Vision}, 2016.

\bibitem[Wetzstein et~al.(2012)Wetzstein, Lanman, Hirsch, and
  Raskar]{wetzstein2012tensor}
G. Wetzstein, D. Lanman, M. Hirsch, and R. Raskar.
\newblock Tensor displays: Compressive light field synthesis using multilayer
  displays with directional backlighting.
\newblock \emph{ACM Transactions on Graphics}, 31\penalty0 (4):\penalty0 1--11,
  2012.

\bibitem[Wilburn et~al.(2005)Wilburn, Joshi, Vaish, Talvala, Antunez, Barth,
  Adams, Horowitz, and Levoy]{wilburn2005high}
Bennett Wilburn, Neel Joshi, Vaibhav Vaish, Eino-Ville Talvala, Emilio Antunez,
  Adam Barth, Andrew Adams, Mark Horowitz, and Marc Levoy.
\newblock High performance imaging using large camera arrays.
\newblock \emph{ACM Transactions on Graphics}, 24\penalty0 (3):\penalty0
  765--776, 2005.

\bibitem[Wu et~al.(2019)Wu, Boominathan, Chen, Sankaranarayanan, and
  Veeraraghavan]{Wu_2019_ICCP}
Yicheng Wu, Vivek Boominathan, Huaijin Chen, Aswin Sankaranarayanan, and Ashok
  Veeraraghavan.
\newblock Phasecam3{D} ― learning phase masks for passive single view depth
  estimation.
\newblock In \emph{International Conference on Computational Photography},
  pages 1--12, 2019.

\bibitem[Xu et~al.(2022)Xu, Jiang, Wang, Fan, Shi, and Wang]{Xu_2022_SinNeRF}
Dejia Xu, Yifan Jiang, Peihao Wang, Zhiwen Fan, Humphrey Shi, and Zhangyang
  Wang.
\newblock {SinNeRF}: Training neural radiance fields on complex scenes from a
  single image.
\newblock In \emph{European Conference on Computer Vision}, 2022.

\bibitem[Yoshida et~al.(2018)Yoshida, Torii, Okutomi, Endo, Sugiyama,
  Taniguchi, and Nagahara]{Yoshida_2018_ECCV}
Michitaka Yoshida, Akihiko Torii, Masatoshi Okutomi, Kenta Endo, Yukinobu
  Sugiyama, Rin-ichiro Taniguchi, and Hajime Nagahara.
\newblock Joint optimization for compressive video sensing and reconstruction
  under hardware constraints.
\newblock In \emph{European Conference on Computer Vision}, 2018.

\bibitem[Yu et~al.(2021)Yu, Ye, Tancik, and Kanazawa]{yu2021pixelnerf}
Alex Yu, Vickie Ye, Matthew Tancik, and Angjoo Kanazawa.
\newblock {pixelNeRF}: Neural radiance fields from one or few images.
\newblock In \emph{Proceedings of the IEEE/CVF Conference on Computer Vision
  and Pattern Recognition (CVPR)}, 2021.

\bibitem[Zhou et~al.(2018)Zhou, Gallego, Rebecq, Kneip, Li, and
  Scaramuzza]{Zhou2018Semi}
Yi Zhou, Guillermo Gallego, Henri Rebecq, Laurent Kneip, Hongdong Li, and
  Davide Scaramuzza.
\newblock Semi-dense {3D} reconstruction with a stereo event camera.
\newblock In \emph{European Conference on Computer Vision}, pages 242--258,
  2018.

\bibitem[Zhou et~al.(2021)Zhou, Gallego, and Shen]{Zhou2021Event}
Yi Zhou, Guillermo Gallego, and Shaojie Shen.
\newblock Event-based stereo visual odometry.
\newblock \emph{IEEE Transactions on Robotics}, 37\penalty0 (5):\penalty0
  1433--1450, 2021.

\end{thebibliography}
}

\end{document}